\documentclass[fullpaper,final]{nldl}
\paperID{9}
\vol{233}
\usepackage{mathtools}

\usepackage{graphicx}

\usepackage{booktabs}

\usepackage{enumitem}

\usepackage{algorithm}
\usepackage{algorithmic}

\usepackage{listings}
\usepackage{hyperref}
\usepackage{url}
\hypersetup{
  pdfusetitle,
  colorlinks,
  linkcolor = BrickRed,
  citecolor = NavyBlue,
  urlcolor  = Magenta!80!black,
}

\usepackage{authblk}
\usepackage{amsmath,amssymb,amsfonts}
\usepackage{latexsym}
\usepackage{textcomp}
\usepackage{xcolor}
\usepackage{arydshln}
\usepackage{adjustbox}
\usepackage{nicefrac}       % compact symbols for 1/2, etc.
\usepackage{float}
\usepackage{gensymb}
\usepackage{times}
\usepackage{soul}
\usepackage{subcaption}
\usepackage{tikz}
\usepackage{pgfplots}

\newcommand{\NAME}{PPDL}
\newcommand{\NAMEVAR}{PPDL-var}
\pgfplotsset{compat=1.18}

\title{Efficient Node Selection in Private Personalized Decentralized Learning}
\author[1,2]{Edvin Listo Zec\thanks{Corresponding Author.}}
\author[3]{Johan Östman}
\author[1]{Olof Mogren}
\author[3]{Daniel Gillblad}
\affil[1]{RISE Research Institutes of Sweden}
\affil[2]{KTH Royal Institute of Technology}
\affil[3]{AI Sweden}
\affil[ ]{\texttt{edvin.listo.zec@ri.se}}

\begin{document}
\maketitle

\begin{abstract}
Personalized decentralized learning is a promising paradigm for distributed learning, enabling each node to train a local model on its own data and collaborate with other nodes to improve without sharing any data. However, this approach poses significant privacy risks, as nodes may inadvertently disclose sensitive information about their data or preferences through their collaboration choices. In this paper, we propose Private Personalized Decentralized Learning (\textbf{\NAME{}}), a novel approach that combines secure aggregation and correlated adversarial multi-armed bandit optimization to protect node privacy while facilitating efficient node selection. By leveraging dependencies between different arms, represented by potential collaborators, we demonstrate that \NAME{} can effectively identify suitable collaborators solely based on aggregated models. Additionally, we show that \NAME{} surpasses previous non-private methods in model performance on standard benchmarks under label and covariate shift scenarios.
\end{abstract}

\section{Introduction}
\label{intro}

Collaborative machine learning is a recent paradigm where multiple actors train a joint model without revealing their local datasets~\cite{mcmahan2017fl}.
Instead, only the locally trained model parameters are shared among the actors.
In applications pertaining to sensitive data, e.g., healthcare and banking, where it may be challenging to collect the data in a single location, collaborative learning has the potential to unlock a plethora of novel collaborations. Collaborative learning is typically distinguished with regard to the underlying network topology.
To this end, federated learning (FL) refers to a star topology where an orchestrating parameter server receives model updates from the actors, aggregates the updates, and broadcasts the aggregate.
Decentralized learning (DL) constitutes arbitrary network topologies without an orchestrator, i.e., actors in the network learn by exchanging model updates within their neighborhood~\cite{lian2017decentral}.
Actors within DL are typically referred to as nodes.

There are inherent risks and limitations with FL, such as that it may be challenging to find a trustworthy third party due to regulations or the desire for autonomy (e.g. for hospitals, banks, or other big corporations). 
Further, FL scales poorly in the number of nodes due to the communication bottleneck and the server constitutes a single-point-of-failure~\cite{lian2017decentral}.
This has motivated research on fully decentralized systems, which eliminate the need for a central server. 
Instead, model parameters are directly communicated between peers in the learning setup using a communication protocol, such as gossip learning~\cite{kempe2003gossip}. 
However, this approach is not well-suited for non-iid settings, where multiple distinct learning objectives may be present. 
In such cases, node selection during training is crucial for achieving efficient and effective learning.

The idea of each node identifying useful peers in the network to train a personalized model was proposed in~\cite{zantedeschi2020collaboration}.
Therein, nodes jointly learn a collaboration graph, via an alternating optimization method, that dictates whom to communicate to.
A score-based method, decentralized adaptive clustering (DAC), was presented in~\cite{zec2022decentralized} where each node scores its neighboring peers based on the the empirical loss, obtained by evaluating the received model parameters on the local dataset.
While DAC manages to find beneficial nodes and identifies heterogeneous clusters in the network, model parameters from the nodes' training updates are still communicated over the network in plain text and the peers receiving the updates must hence be trusted. As such, DAC is vulnerable to inference attacks. This raises the question of how to ensure the privacy of the model parameters in decentralized machine learning systems. 
In many privacy-critical applications, differential privacy~\cite{dwork2006calibrating} is used in conjunction with FL to protect the data of nodes. 
Although this adds a layer of privacy, it comes at the expense of a deterioration in model performance.

In this work, we overcome this problem and introduce a communication-efficient and privacy-preserving algorithm named \textbf{Private Personalized Decentralized Learning (\NAME{})}. We use multi-armed bandits to find beneficial collaborators and secure aggregation~\cite{bonawitz2017secagg,tjell2021secagg} to hide individual updates. Our method works in a server-less decentralized setting, but can also apply to standard FL. We protect against inference attacks by only observing aggregated models.In our proposed method, a peer only observes an aggregate of model parameters, which substantially lessens the risk of inference attacks as compared to previous works.

Since a node only receives an aggregate of the parameter updates of $M$ nodes at a given point in time, it cannot infer a score on the similarity of any one of the peers in the aggregate (as in DAC); such a score can only be computed for the aggregate. 
Instead, our solution exploits dependencies between different group selections and makes use of adversarial multi-armed bandit optimization to efficiently find the subsets of peers that are beneficial for collaboration. Our experimental evaluations demonstrate that our approach offers a competitive solution for personalized decentralized learning that preserves data privacy under covariate shift and label shift and efficiently finds the beneficial collaborators within the network. Our solution has a communication efficiency and performance similar to that of previous methods, but adds a higher level of privacy.

\section{Decentralized learning by finding useful collaborations}
\textbf{Problem formulation.} We consider several DL tasks over a network of $K$ nodes, each with a \textit{private} data distribution $\mathcal{D}_i$ over the inputs $x\in\mathcal{X}$ and labels $y\in\mathcal{Y}$.
Each node $i\in [K]$ has a model $f_i$ with parameters $w_i\in \mathbb{R}^d$ and a loss function $\ell(f_i(w_i;x), y): \mathbb{R}^d \times \mathcal{X} \times \mathcal{Y} \rightarrow \mathbb{R}$.
Each note wants to minimize its expected loss over its data,
\begin{equation}\label{eq:objective}
    w^\star_i = \arg\min_{w_i\in\mathbb{R}^d} \mathbb{E}_{(x,y)\sim \mathcal{D}_i}\left[ \ell(f_i(w_i;x), y)\right].
\end{equation}
A challenge is to find similar nodes to collaborate with, without sharing data. If the distributions are substantially dissimilar, collaboration may result in decreased performance compared to local training without collaboration. In situations where some of the other nodes in the network have similar local data distributions, it may be beneficial to collaborate towards the goal in~\eqref{eq:objective} by means of exchanging and aggregating model parameters. 

However, revealing details of node data may be difficult or impossible due to privacy reasons. To address this issue, we propose a method for identifying nodes with similar local datasets in a private manner. We assume the nodes communicate over a network $\mathcal{G}=(\mathcal{N}, \mathcal{E})$ where $\mathcal{N}=\lbrace 1,\dots,K \rbrace$ are the nodes and $\mathcal{E}=\lbrace (i,j):i, j\in \mathcal{N}, i\neq j\rbrace $ are the edges between the nodes. 
The neighborhood of node $i\in\mathcal{N}$ is denoted by $\mathcal{N}_i = \lbrace j: (i,j) \in \mathcal{E}, j\in\mathcal{N}  \rbrace $.
Like~\cite{zec2022decentralized,sui2022friends}, node $i$ want to find a set of nodes $\mathcal{M}_i\subseteq \mathcal{N}_i$ to exchange models with. In each round, the learning proceeds as follows. First, each node $i\in \mathcal{N}$ selects a set $\mathcal{M}_i \subseteq \mathcal{N}_i$ to receive model updates from. Second, the nodes in $\mathcal{M}_i$ submit their local models securely to node $i$ by using secure aggregation, e.g.,~\cite{tjell2021secagg}. Third, node $i$ computes the aggregated model from the nodes in $\mathcal{M}_i$ and aggregates it with its local model after which local training is initiated using the updated model.

\textbf{Privacy.} Although FedAvg is commonly advertised as being private, recent results have demonstrated attacks able to recover training data from the models~\cite{dimitar2022leakage}.
To protect the nodes from such attacks, we utilize secure aggregation to ensure that a node who queried multiple model parameters from a subset of its neighbors only get to observe an aggregate of those models.
The design of secure aggregation schemes is outside of the scope of this work but may be achieved for arbitrary networks by using Shamir's secret sharing scheme~\cite{shamir1979secret} as demonstrated in~\cite{tjell2021secagg}.
For our purposes, we assume that a node $i$ queries a set $\mathcal{M}_i^{(t)} \subseteq \mathcal{N}_i$ of size $M$ in round $t\in [T]$ and observes only the aggregate $\bar{w}_i^{(t)}=\sum_{j\in\mathcal{M}_i^{(t)}} \beta_j w_j$ where $\beta_j\geq0$ satisfy $\sum_{j\in \mathcal{M}_i^{(t)}}\beta_j=1$.
Consequently, node $i$ is presented with $C_i = {\vert \mathcal{N}_i \vert \choose M}$ different groups of nodes to choose among where we assume $\vert \mathcal{N}_i \vert \geq M$ for all $i\in[N]$.
For example, in a fully connected network consisting of $K=100$ nodes and secure aggregation schemes where $M=2$ and $M=3$, we have 4,851 and 156,849 different groups, respectively.

\textbf{Multi-armed bandits.} We have a challenging group-selection problem with many groups and few rounds. A node can only evaluate a group by its local accuracy, which is stochastic and non-stationary due to other nodes’ actions. We use an online learning approach and model the problem for each node as an adversarial multi-armed bandit with $C_i$ arms and $T$ rounds~\cite{auer02adversarial}.

The performance of a bandit algorithm is measured by pseudo-regret, which compares the expected rewards of the best arm and the algorithm. For adversarial bandits, the pseudo-regret per round decreases as $\mathcal{O}(\sqrt{C_i/T})$~\cite{audibert2009minimax}.
This means a large $C_i$, as in our case, an algorithm cannot be expected to perform well in a few rounds.
However, this assumes independent rewards; if rewards are dependent, pulling an arm can give information about other arms and reduce exploration~\cite{gupta2021correlated}.

In our problem, some groups share nodes. The number of groups that share $u$ nodes with a given group is ${M \choose u } {N-M-1 \choose M-u}$. For example, in a fully connected network with $N=100$ and $M=3$, there are 13,680 and 288 groups that share one and two nodes, respectively, with a given group. So, selecting one group out of the 156,849 could inform about 13,968 groups. To leverage this idea, we use of pseudo-rewards, as presented in~\cite{gupta2021correlated}. 

Let the different groups available to node $i$ be indexed from $1,\dots,C_i$ and, w.l.o.g., let the reward from choosing group $j\in[C_i]$ at time $t$ satisfy $r_j^{(t)}\in [0,1]$. 
We define the pseudo-rewards $s_{l,j}^{(t)}(\alpha_j^{(t)}) \in [0,1]$ as an upper bound on the expected reward on $r_l^{(t)}$ given that we observe $r_j^{(t)}$ for $j\in [C_i]$ and $l\in [C_i]\setminus \lbrace j \rbrace$.
This is mathematically represented as:
\begin{equation}\label{eq:pseudoreward}
\mathbb{E}\left[ r_l^{(t)} \vert r_j^{(t)} = \alpha_j^{(t)} \right] \leq s_{l,j}^{(t)}(\alpha_j^{(t)}).
\end{equation}
For $j=l$, we let $s_{j,j}^{(t)} = r_j^{(t)}$.
Note that setting $s_{l,j}^{(t)}(\alpha_j^{(t)})=1$ for all $j,l\in [C_i]$, $l\neq j$ and $t\in[T]$, results in recovering the uncorrelated multi-armed bandit setting.
Note that the inequality in~\eqref{eq:pseudoreward} must be satisfied in order to achieve zero-regret asymptotically in the number of rounds~\cite{gupta2021correlated}. However, as our objective is to simply identify nodes with similar local data distributions within a fixed number of training rounds, the choice of pseudo-reward in~\eqref{eq:pseudoreward} will mainly serve to trade-off between exploitation and exploration.

To use the correlated bandit framework in our setting, we notice that groups with large overlap have many parameters in common in the aggregation step, hence, it seems plausible that also their expected rewards should be closer than groups with less overlap.
Therefore, we design the pseudo-rewards between two groups to be decreasing in the number of overlapping nodes.
Furthermore, it is expected that the discrepancy in accuracy between groups with large overlap decreases over time, hence the time dependency in~\eqref{eq:pseudoreward}.
Let $u_{l,j}\in \lbrace0, \dots, M-1\rbrace$ denote the number of overlapping nodes between group $l$ and group $j$.
We consider pseudo-rewards of the form
\begin{equation}\label{eq:ub}
    s_{l,j}^{(t)}(\alpha_j^{(t)}) = \min\left\lbrace\alpha_j^{(t)} + \frac{q(t)}{u_{l,j}}, 1\right\rbrace 
\end{equation}
where $q: [T] \rightarrow \mathbb{R}_+$ is a non-increasing function in time, i.e., $q(t_2)\leq q(t_1)$ for $t_2> t_1$.
We make this choice as the variance between node models is anticipated to decrease as models converge.

\subsection{Private multi-armed bandits for node selection}
In this section, we present our bandit algorithm for a node. For ease of notation, we omit the node index. 
Let $k^{(t)}\in [C_i]$ be the group chosen at time $t$ and let $n_{k^{(t)}}(t)$ be the number of times it has been chosen.
The reward from choosing group $j\in[C_i]$ is defined as
$\mu_j(t) = \frac{\sum_{\tau=1}^t \mathbf{1}\lbrace k^{(\tau)} = j \rbrace  r_j^{(\tau)}}{n_j(t)}$
and the pseudo-reward for group $l\in [C_i] \setminus \lbrace j \rbrace$ when group $j\in [C_i]$ is selected, is given by
$\phi_{l,j}(t) = \frac{\sum_{\tau=1}^t \mathbf{1}\lbrace k^{(\tau)} = j \rbrace  s_{l,j}^{(\tau)}(r_j^{(\tau)})}{n_j(t)}$.
We reduce the problem size by selecting only competitive arms, i.e., arms whose minimum pseudo-rewards are higher than the maximum reward.  
To this end, we define the set of significant arms as $\mathcal{S}_i^{(t)} = \lbrace j \in [C_i] : n_j(t) > t/N \rbrace$ and let $\bar{k}^{(t)} = \arg\max_{l\in \mathcal{S}_i^{(t)}}\mu_l(t)$.
The set of empirically competitive arms is defined as
\begin{equation}\label{eq:comp_set}
    \mathcal{A}_i^{(t)} =\left\lbrace j\in [C_i] : \min_{l \in \mathcal{S}_i^{(t)}} \phi_{j,l}(t) \geq \mu_{\bar{k}^{(t)}}(t)  \right\rbrace \cup \lbrace \bar{k}^{(t)} \rbrace.
\end{equation}
Note that $\mathcal{A}_i^{(t)}$ is not monotonically decreasing in $t$ as arms may be non-competitive in one round and competitive in the next.
Once $\mathcal{A}_i^{(t)}$ has been obtained, an arbitrary multi-armed bandit algorithm may be applied over the set of arms. As we consider adversarial rewards, we employ the \textit{Tsallis-Inf} algorithm that is known to achieve a pseudo-regret with the optimal scaling~\cite{zimmert2021tsallis}, where large $q(t)$ encourages exploration whereas a small $q(t)$ encourages exploitation.
%The algorithm is based on online mirror descent with Tsallis entropy regularization, the reader is referred to Section~3 in~\cite{zimmert2021tsallis} for further details. Note that $q(t)$ in~\eqref{eq:ub} has an important role in~\eqref{eq:comp_set} as it dictates the size of the empirical pseudo-rewards. 

%\subsection{Complexity}

%A naive implementation of \NAME{} would require one to store the empirical pseudo rewards of all arms for each unique action.
%Hence, for node $i\in\mathcal{N}_i$, the worst-case storage requirement applies to a fully connected network and scales as $\mathcal{O}(T C_i)$ over the number of training rounds.
%Such scaling may be acceptable for small networks, say $N\leq100$ and limited privacy, e.g., $M\leq 3$.
%To go beyond and allow for arbitrary network- and group sizes, one may evaluate the empirical pseudo-rewards on demand and, further, approximate the policy update by updating only the most likely actions in Algorithm~\ref{alg:bandit}.

\section{Experiments}
Our code is made available upon publication to encourage reproducibility \footnote{\url{https://github.com/edvinli/ppdl}}. All experiments were carried out on an Nvidia 3090 Ti GPU. We conduct experiments on various cluster configurations and employ the CIFAR-10 and Fashion-MNIST datasets, which are commonly used in the literature for decentralized machine learning evaluations on covariate and label shift, see Section~3.1~\cite{kairouz2019advances}. 
We follow previous work \cite{zec2022decentralized,sui2022friends} and assume a fully connected graph among the nodes. Our algorithm aims to find a sub-graph for each node that maximizes its local task performance. In other words, we want to find the best collaborators for each node based on its local, private data.

\textbf{Baselines.} In all experiments we use decentralized adaptive clustering (\textbf{DAC}) \cite{zec2022decentralized} as a baseline for comparison, as it is most similar to our work. 
In addition, we also make comparisons to a random gossip communication protocol (denoted \textbf{Random}) and an oracle (denoted \textbf{Oracle}) that has perfect information of cluster assignments and only communicates (randomly) within these. 
Moreover, we also compare with local training on the nodes where no communication is allowed (denoted \textbf{Local}).

\textbf{Covariate shift.} To evaluate the performance of our method under non-iid data distributions, we replicate some of the experiments outlined in \cite{zec2022decentralized} for covariate shift with 100 nodes by dividing the data uniformly into four partitions, each with images rotated $0\degree, 90 \degree, 180\degree$ and $270 \degree$, respectively. We also experiment with heterogeneous cluster sizes by dividing the data into clusters of $0\degree, 180 \degree, 350\degree$ and $10 \degree$ rotation, with $70, 20, 5$ and $5$ nodes in each cluster, respectively.

\textbf{Label shift.} Moreover, we also conduct experiments on label shift. As in \cite{zec2022decentralized}, for the CIFAR-10 dataset we divide the data into two clusters based on labels, one for animal images and one for vehicle images. Additionally, we extend the experiments on label shift where we partition the data such that each node only has two labels, and these labels are grouped into clusters of five, where each cluster contains 20 nodes with the same two labels.

In our experiments, we evaluate all models on a test set with the same distributional shift as the training set for each node in the network. This is because the goal is to solve the local learning task for each node as effectively as possible. We use early stopping locally on each node.

\textbf{Model and data.} We use the same CNN architecture as \cite{zec2022decentralized}, with three convolutional and two fully connected layers. We simulate 100 nodes for CIFAR-10 and Fashion-MNIST, and average results over three runs. Each node has equal data samples, uses the Adam optimizer and batch size of 8, and samples $M=3$ other nodes per round. We train for three local epochs and 200 rounds. We use two $q(t)$ in~\eqref{eq:ub}: constant (\NAME{}) and exponentially decaying (\NAMEVAR{}), tuned by validation. We also tune learning rates using a validation set.
%For the covariate shift problem, we used a learning rate of $\eta = 3\cdot 10^{-4}$, and for the label shift problem, we used a learning rate of $\eta = 3\cdot 10^{-5}$. 
%For the DAC baseline we use a value of $\tau=30$ which is the same as reported in the paper \cite{zec2022decentralized}. 

\begin{figure*}
    \centering
    \begin{subfigure}[t]{0.33\linewidth}
    \includegraphics[width=\columnwidth]{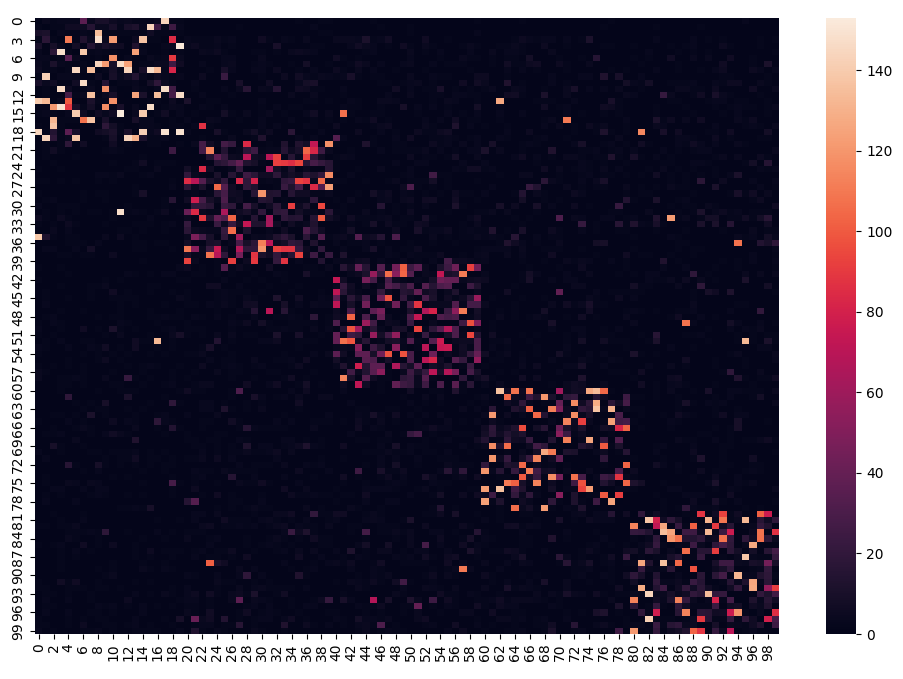}
    \subcaption{\NAME{}}
    \end{subfigure}
    \begin{subfigure}[t]{0.33\linewidth}
    \includegraphics[width=\columnwidth]{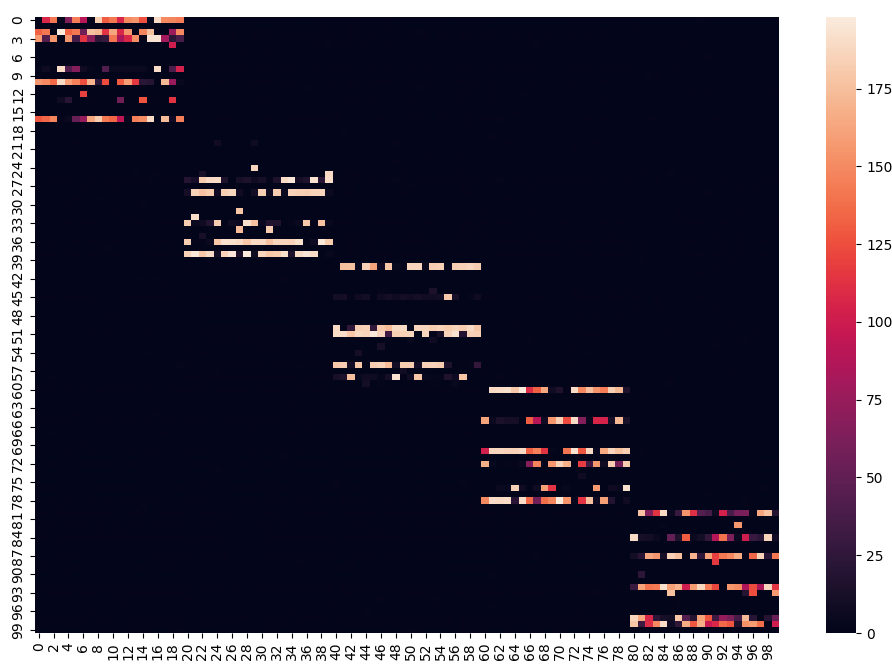}
    \subcaption{DAC}
    \end{subfigure}
    \begin{subfigure}[t]{0.33\linewidth}
    \includegraphics[width=\columnwidth]{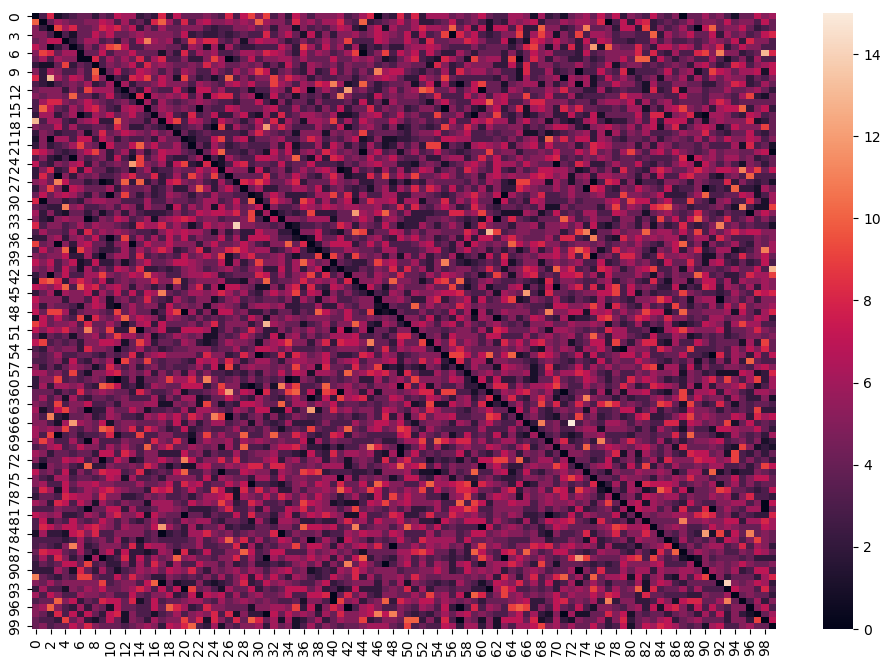}
    \subcaption{Random}
    \end{subfigure}
    \begin{subfigure}[t]{0.33\linewidth}
    \includegraphics[width=\columnwidth]{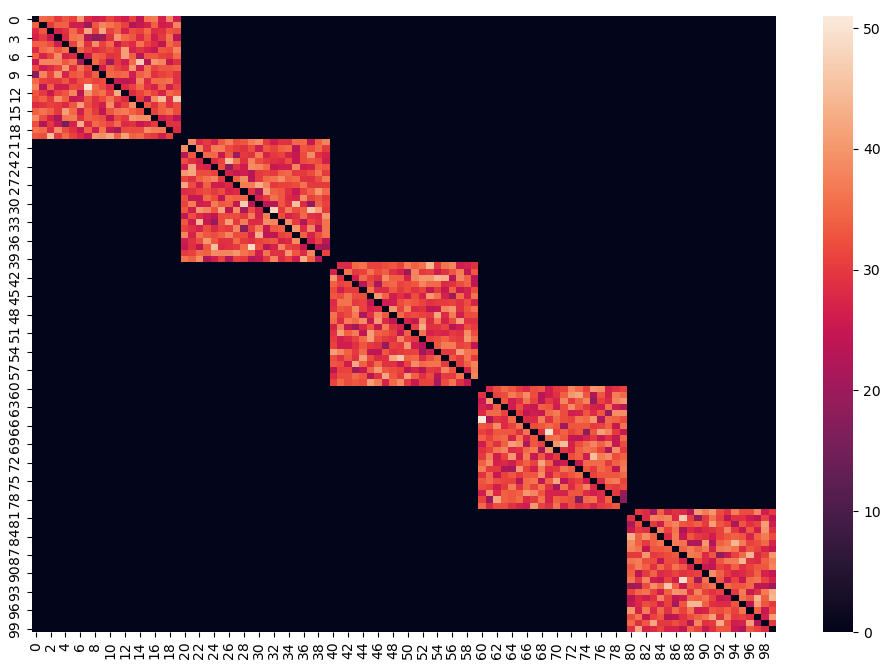}
    \subcaption{Oracle}
    \end{subfigure}
    \caption{Heatmaps visualising how often node $x$ communicates with node $y$ for the four different methods on the CIFAR-10 dataset with 5 clusters.}
    \label{fig:heatmap}
\end{figure*}

\begin{table*}[t]
\centering
\caption{CIFAR-10 label shift test accuracy with 5 clusters.}
\label{tab:label3}
\begin{tabular}{lcccccc}
\toprule
Method  & Cluster 0 &  Cluster 1 & Cluster 2 & Cluster 3 & Cluster 4 &  Mean \\
\midrule
\NAME{} & 74.70 & 68.51 & 73.78 & 77.74 & 74.60 & 73.87 \\
\NAMEVAR & 71.82 & 80.40 & 78.72 & 82.06 & 76.05 & 77.81 \\
\textbf{DAC} & \textbf{79.41} & \textbf{76.83} & \textbf{78.52} & \textbf{80.58} & \textbf{76.94} & \textbf{78.46} \\
Random & 68.90 & 63.31 & 66.70 & 69.98 & 69.92 & 67.76 \\
Local & 77.11 & 88.79 &  82.60 & 61.83 & 68.32 & 75.73\\ \hdashline
\textcolor{gray}{Oracle} & \textcolor{gray}{88.65} & \textcolor{gray}{91.25} & \textcolor{gray}{84.69} & \textcolor{gray}{81.54} & \textcolor{gray}{79.62} & \textcolor{gray}{85.15} \\
\bottomrule
\end{tabular}
\end{table*}

\begin{table}
\centering
\caption{Test accuracies for covariate shift on CIFAR-10 and Fashion-MNIST, with varying node numbers per cluster (70,20,5,5). Mean values over clusters are also provided.}
\label{tab:covariate180}
\begin{tabular}{lcccccc}
\toprule
\multicolumn{7}{c}{\textbf{CIFAR-10}}\\
Method  &  $0\degree$ & $180\degree$ & $350\degree$ & $10\degree$ &  Mean \\
\midrule
\NAME{} & 52.37 & 45.21 & 50.60 & 51.03 & 49.80 \\
\textbf{\NAMEVAR{}} & \textbf{54.63} & \textbf{47.27} & \textbf{51.84} & \textbf{53.30} & \textbf{51.76} \\
DAC & 53.70 & 47.73 & 52.84 & 51.35 & 51.41 \\
Random & 54.85 & 44.70 & 52.64 & 52.43 & 51.16 \\
Local & 34.06 & 31.64  & 29.92 & 32.68 & 31.91 \\ \hdashline
\textcolor{gray}{Oracle} & \textcolor{gray}{55.04} & \textcolor{gray}{46.80} & \textcolor{gray}{38.35} & \textcolor{gray}{38.00} & \textcolor{gray}{44.55} \\
\bottomrule
\multicolumn{7}{c}{\textbf{Fashion-MNIST}}\\
\textbf{\NAME{}} & \textbf{84.62} & \textbf{81.81} & \textbf{81.01} & \textbf{82.11} & \textbf{82.39} \\
\NAMEVAR{} & 80.68 & 81.12 & 80.42 & 80.66 & 80.72 \\
DAC &  82.48 & 80.44 & 79.85 & 80.43 & 80.80 \\
Random & 84.26 & 79.61 & 78.42 & 78.99 & 80.32\\
Local & 78.72 & 76.83 & 77.40 & 77.26 & 77.55\\ \hdashline
\textcolor{gray}{Oracle} & \textcolor{gray}{83.00} & \textcolor{gray}{81.93} & \textcolor{gray}{79.01} & \textcolor{gray}{79.76} & \textcolor{gray}{80.93}\\
\bottomrule
\end{tabular}
\end{table}

\begin{table}
\centering
\caption{Test accuracies for covariate shift on CIFAR-10 and Fashion-MNIST, with the same number of nodes per cluster (25). Mean values over clusters are also provided.}
\label{tab:covariate90}
\begin{tabular}{lcccccc}
\toprule
\multicolumn{7}{c}{\textbf{CIFAR-10}}\\
Method  &  $0\degree$ & $90\degree$ & $180\degree$ & $270\degree$ &  Mean \\
\midrule
\NAME{} & 43.48 & 43.31 & 43.73 & 43.10 & 43.40 \\
\NAMEVAR{} & 45.06 & 44.05 & 44.60 & 43.14 & 44.22 \\
\textbf{DAC} & \textbf{45.21} & \textbf{45.08} & \textbf{45.18} & \textbf{45.78} & \textbf{45.31} \\
Random & 41.35 & 41.19 & 42.39 & 41.46 & 41.60 \\
Local & 32.01 & 32.34 & 31.47 & 33.07 & 32.22 \\ \hdashline
\textcolor{gray}{Oracle} & \textcolor{gray}{49.47} & \textcolor{gray}{49.66} & \textcolor{gray}{49.57} & \textcolor{gray}{48.43} & \textcolor{gray}{49.28} \\
\bottomrule
\multicolumn{7}{c}{\textbf{Fashion-MNIST}}\\
\NAME{} & 80.69 & 81.12 & 80.43 & 80.66 & 80.73 \\
\textbf{\NAMEVAR{}} & \textbf{80.81} & \textbf{81.71} & \textbf{82.36} & \textbf{80.19} & \textbf{81.26} \\
DAC & 78.83 & 79.51 & 78.69 & 79.02 & 79.01\\ 
Random & 80.20 & 80.72 & 79.3 & 79.99 & 80.05 \\ 
Local & 78.84 & 79.36 & 79.98 & 77.04 & 78.81 \\\hdashline
\textcolor{gray}{Oracle} & \textcolor{gray}{82.86} & \textcolor{gray}{83.18} & \textcolor{gray}{84.25} & \textcolor{gray}{83.79} & \textcolor{gray}{83.52} \\
\bottomrule
\end{tabular}
\end{table}
\newpage
\begin{table}
\centering
\caption{CIFAR-10 label shift test accuracy with 'animal' and 'vehicle' clusters.}
\label{tab:label}
\begin{tabular}{lcccc}
\toprule
Method  &  Vehicles & Animals &  Mean \\
\midrule
\NAME{} & $51.86$ & $36.31$ & $43.81$ \\
\textbf{\NAMEVAR{}} & $\textbf{52.86}$ & $\textbf{36.33}$ & $\textbf{44.60}$ \\
DAC & $52.78$ &  $33.87$ &  $43.32$ \\
Random  &  $44.79$ &  $30.00$ &  $37.40$ \\
Local & $51.10$ & $35.11$ & $43.11$ \\ \hdashline
\textcolor{gray}{Oracle} &  \textcolor{gray}{57.17} & \textcolor{gray}{39.74}  & $\textcolor{gray}{48.45}$ \\
\bottomrule
\end{tabular}
\end{table}

\textbf{Covariate shift.} Tables~\ref{tab:covariate180}~and~\ref{tab:covariate90} show the results of our covariate shift experiments with two cluster setups. Our method, \NAME{}, performs similarly to DAC, but with secure aggregation for privacy. Random favors large clusters and penalizes small ones, as seen in Table~\ref{tab:covariate180}. DAC and \NAME{} avoid collaborating with “poisonous” nodes by their sampling schemes, improving test accuracy in the $180\degree$ cluster. Oracle has low test accuracies for small clusters, likely due to limited data (only 5 nodes per cluster). For the smallest clusters, $350\degree$ and $10\degree$, \NAME{} and DAC find similar nodes in the large $0\degree$ cluster, improving performance. Thus, DAC and \NAME{} increase performance and fairness for smaller clusters that differ from large ones. The Fashion-MNIST results are less different between methods, likely due to the easier problem than CIFAR-10. Also, rotating images may not be challenging for small CNNs, as they can learn rotation-invariant representations with enough data. We analyze harder label shift problems ne

\textbf{Label shift.} The results of our label shift experiment with two clusters (animals and vehicles) are presented in Table~\ref{tab:label}. We observe that both \NAME{} and DAC perform well, with \NAME{} achieving superior results. The highest accuracy is achieved with \NAMEVAR{}, in which $q(t)$ is exponentially decayed. We note that Random performs worse than local training without collaboration, likely due to model poisoning caused by nodes communicating with incorrect clusters. For Random, the node models learn different representations for the different clusters, and when merging models from two distinct clusters, the resulting model is inferior due to the significant dissimilarity between the models, a phenomenon known as \textit{client drift.} Both DAC and \NAME{} are able to mitigate this problem by identifying useful collaborators.

The results of our five-cluster experiment on CIFAR-10 are presented in Table~\ref{tab:label3}, where each cluster consists of two unique labels. We observe that Random performs worse than the Local baseline on average also in this setting. Our experiments also reveal a high degree of variance within a cluster for the Local baseline, which can be attributed to the small size of node data. In contrast, the \NAME{} and DAC methods perform comparably and are able to correctly identify beneficial collaborators, as depicted in Figure \ref{fig:heatmap}.

\section{Related work}
%\subsection{Decentralized learning}
\textbf{Decentralized learning.} Previous studies have demonstrated the effectiveness of gossip algorithms, as highlighted in references such as \cite{kempe2003gossip,boyd2006randomized,ormandi2013gossip}. Furthermore, collaborative gossip algorithms, where nodes possess distinct local tasks, have been investigated in the context of multi-task learning (MTL) as seen in \cite{vanhaesebrouck17collab,zantedeschi2020collaboration}. While gossip learning has been demonstrated to be effective in convex optimization, its application in non-convex optimization, which is required for training deep neural networks, has not been as extensively studied. One of the first works that explored the use of gossip-based optimization for non-convex deep learning was conducted on convolutional neural networks (CNNs) in \cite{blot2016gossip}. The authors demonstrated that high accuracies could be achieved at low communication costs using a decentralized and asynchronous framework. However, it is important to note that gossip learning is not well-suited for non-iid settings, where several distinct learning objectives may be present. Indeed, a protocol based on random communication between nodes does not take into consideration the benefits of node selection during training.

In centralized FL, methods based on hard clustering \cite{ghosh2020efficient,mansour2020three,sattler2020clustered} can efficiently identify node clusters, but they limit the collaboration of nodes to their own clusters. This prevents nodes from utilizing useful information from similar clusters in forming a global model. Recent works have advanced decentralized learning of deep neural networks on non-iid data. \cite{sui2022friends} used expectation-maximization, while \cite{ma2022attract} improved node selection and communication cost with gradient-based cosine similarity and model pruning. \cite{onoszko2021decentralized} identified similar nodes by empirical loss, but only allowed hard clustering. \cite{zec2022decentralized} proposed a decentralized adaptive clustering algorithm that used empirical loss similarity to discover beneficial peers, but without privacy guarantees for model weights. This probability vector is then used for sampling similar nodes in the next communication round for each node, allowing for soft cluster assignments and communication within the entire graph. Empirical results demonstrate the effectiveness of this method in identifying clusters of nodes and improving the performance of the models. Although this method identifies useful node collaborations, there are a lot of privacy risks as model weights are being shared without any privacy guarantees.

\textbf{Secure aggregation.}
Secure aggregation is a method to enhance node privacy in FL by protecting against server inference attacks~\cite{bonawitz2017secagg,bell2020poly}.
The idea relies on random masking of the node models, before uploaded to the server, such that the masks cancel out when models are aggregated.
Extensions based on secret sharing schemes~\cite{shamir1979secret} have been proposed, e.g.,~\cite{s02021lightsecagg}. For decentralized learning, where the communication topology may be arbitrary, only few works have considered privacy. One protocol for secure aggregation over arbitrary networks is presented in~\cite{tjell2021secagg}. Specifically, for node $i$, the scheme consists of two phases: i) node $j\in\mathcal{N}_i$ broadcasts a public key that is used to privately collect shares of a random mask generated by node $l\in \mathcal{N}_i$, ii) node $i$ receives the masked models and the aggregated shares of the random masks at from each node and reconstructs the aggregated masks to recover the aggregated model. Note that all of the above schemes require the models to be mapped to a finite field, an operation that may impact the training. A step towards avoiding this step for secure aggregation over connected graphs was recently proposed in~\cite{tjell2021real}.

\textbf{Multi-armed bandits.}
Random node sampling in FL and DL can be improved by biasing towards nodes’ local losses \cite{cho2022bandit}. Multi-armed bandits for node selection were introduced in~\cite{xia2020multi} with rewards based on node latency and objective to minimize training time. Extensions for model averaging~\cite{kim2020comb}, asynchronous FL~\cite{zhu2022staleness}, and dropout and fairness handling~\cite{huang2022volatile} were proposed. However, multi-armed bandits may perform poorly when the number of arms is large or dependent. Dependency-based clustering \cite{singh2020dependent} and pseudo-reward shrinking \cite{gupta2021correlated} are two methods to exploit dependencies and reduce the number of arms.

\section{Conclusions and future work}
We introduce \textbf{Private Personalized Decentralized Learning} (\textbf{\NAME{}}), a novel privacy-preserving node selection approach for personalized decentralized deep learning based on adversarial multi-armed bandits. Our approach uses secure aggregation to hide individual node metrics and exploits node dependencies to sample groups of collaborators efficiently. To the best of our knowledge, this is the first privacy-preserving node selection scheme for decentralized learning. We show that \textbf{\NAME{}} achieves comparable performance to existing (non-private) techniques on multiple experiments, while also providing privacy protection with secure aggregation.

For future research, it would be interesting to explore aggregation methods for models trained on different datasets in order to enhance the robustness of nodes to merging with other clusters.  Another direction is to understand how privacy is affected by the number of nodes participating in the secure aggregation.
Intuitively, as shown in, e.g., Section V.A in~\cite{tjell2021secagg}, privacy improves with larger group sizes.

\printbibliography

\end{document}